\documentclass[default,iicol]{sn-jnl}
    
\jyear{2023}%

\theoremstyle{thmstyleone}%
\theoremstyle{thmstyletwo}%
\theoremstyle{thmstylethree}%

\begin{document}

\title[Article Title]{Resolution Enhancement Processing on Low Quality Images Using Swin Transformer Based on Interval Dense Connection Strategy}

\author[1]{\fnm{Rui-Yang} \sur{Ju}}\email{jryjry1094791442@gmail.com}
\equalcont{These authors contributed equally to this work.}

\author[1]{\fnm{Chih-Chia} \sur{Chen}}\email{crystal88irene@gmail.com}
\equalcont{These authors contributed equally to this work.}

\author*[1]{\fnm{Jen-Shiun} \sur{Chiang}}\email{jsken.chiang@gmail.com}

\author[1]{\fnm{Yu-Shian} \sur{Lin}}\email{abcpp12383@gmail.com}

\author[1]{\fnm{Wei-Han} \sur{Chen}}\email{kj211378@gmail.com}

\author[1]{\fnm{Chun-Tse} \sur{Chien}}\email{popper0927@hotmail.com}

\affil[1]{\orgdiv{Department of Electrical and Computer Engineering}, \orgname{Tamkang University}, \orgaddress{\street{No.151, Yingzhuan Rd., Tamsui Dist.}, \city{New Taipei City}, \postcode{251301}, \country{Taiwan}}}

\abstract{
The Transformer-based method has demonstrated remarkable performance for image super-resolution in comparison to the method based on the convolutional neural networks (CNNs). However, using the self-attention mechanism like SwinIR (Image Restoration Using Swin Transformer) to extract feature information from images needs a significant amount of computational resources, which limits its application on low computing power platforms. To improve the model feature reuse, this research work proposes the Interval Dense Connection Strategy, which connects different blocks according to the newly designed algorithm. We apply this strategy to SwinIR and present a new model, which named SwinOIR (Object Image Restoration Using Swin Transformer). For image super-resolution, an ablation study is conducted to demonstrate the positive effect of the Interval Dense Connection Strategy on the model performance. Furthermore, we evaluate our model on various popular benchmark datasets, and compare it with other state-of-the-art (SOTA) lightweight models. For example, SwinOIR obtains a PSNR of 26.62 dB for $\times$4 upscaling image super-resolution on Urban100 dataset, which is 0.15 dB higher than the SOTA model SwinIR. For real-life application, this work applies the lastest version of You Only Look Once (YOLOv8) model and the proposed model to perform object detection and real-life image super-resolution on low-quality images. This implementation code is publicly available at \url{https://github.com/Rubbbbbbbbby/SwinOIR}.
}

\maketitle

\section{Introduction}
Image super-resolution is a hot research topic in image processing and computer vision (CV), which involves enhancing low-quality input images into high-quality output images. With the rise of deep learning, neural networks for image super-resolution have started to develop by leaps and bounds \cite{dong2015image}\cite{yue2016image}\cite{wang2020deep}. convolutional neural networks (CNNs) \cite{yang2019deep}\cite{tian2020coarse}\cite{tian2020lightweight}\cite{tian2022heterogeneous}\cite{tian2022image}\cite{tian2022image} have become the main network models for image super-resolution in recent years. Although CNNs improve model performance by designing new network architectures, such as using various algorithms to connect the convolution layers \cite{ju2022threshnet}\cite{ju2023efficient}, this enhancement cannot resolve the problem of the lack of interactive content between convolution kernels and images. The same convolution kernel does not perform well in image super-resolution for different images.

In recent years, Transformers \cite{vaswani2017attention}\cite{han2021transformer} have achieved outstanding success in CV by employing the self-attention mechanism to gather context information. With several proposed Transformer architecture models \cite{yao2023dual}\cite{wang2022pvt}\cite{zhang2023vitaev2} achieving the state-of-the-art (SOTA) performance in dense prediction tasks, researchers have also made significant advancements in image super-resolution \cite{liang2022light}\cite{lei2021transformer}\cite{gao2023ctcnet}\cite{liu2022interactformer}.

In this paper, we introduce a new connection strategy, which is named Interval Dense Connection Strategy, to connect different blocks. This work applies it to SwinIR (Image Restoration Using Swin Transformer) \cite{liang2021swinir}, presenting SwinOIR (Object Image Restoration Using Swin Transformer) for image super-resolution. More specifically, the SwinOIR model consists of three modules: Pre-Feature Extraction, Main Feature Extraction, and High Quality Image Reconstruction. In Main Feature Extraction, it connects different blocks according to the newly designed algorithm to mitigate the gradient disappearance problem of the model, making reverse gradient propagation easier and improving the model convergence.

The contributions of this paper are summarised as follows:
\begin{itemize}
\item [1)]
This research work proposes a new algorithm for the connection of different blocks, and apply this strategy to SwinIR to develop a new model for image super-resolution. Experimental results demonstrate that our model achieves SOTA performance on several popular benchmark datasets.
\end{itemize}
\begin{itemize}
\item [2)]
This paper introduces a two-stage framework and design an application for real-life scenarios, which includes object detection and image super-resolution tasks. Our application makes it easier for users to find details of multiple tiny objects in low-quality images.
\end{itemize}

The rest of this paper is organized as follows: Section \ref{related} describes previous works on Swin Transformer for image super-resolution. The newly proposed model architecture is reported in Section \ref{method}. Section \ref{experiment} presents the experimental results of our model and compares it with other SOTA models. Section \ref{application} introduces our object detection and image super-resolution application for real-life scenarios. Finally, Section \ref{conclusion} discusses this paper's conclusions and future works.

\begin{figure*}[h]
  \centering
  \includegraphics[width=\linewidth]{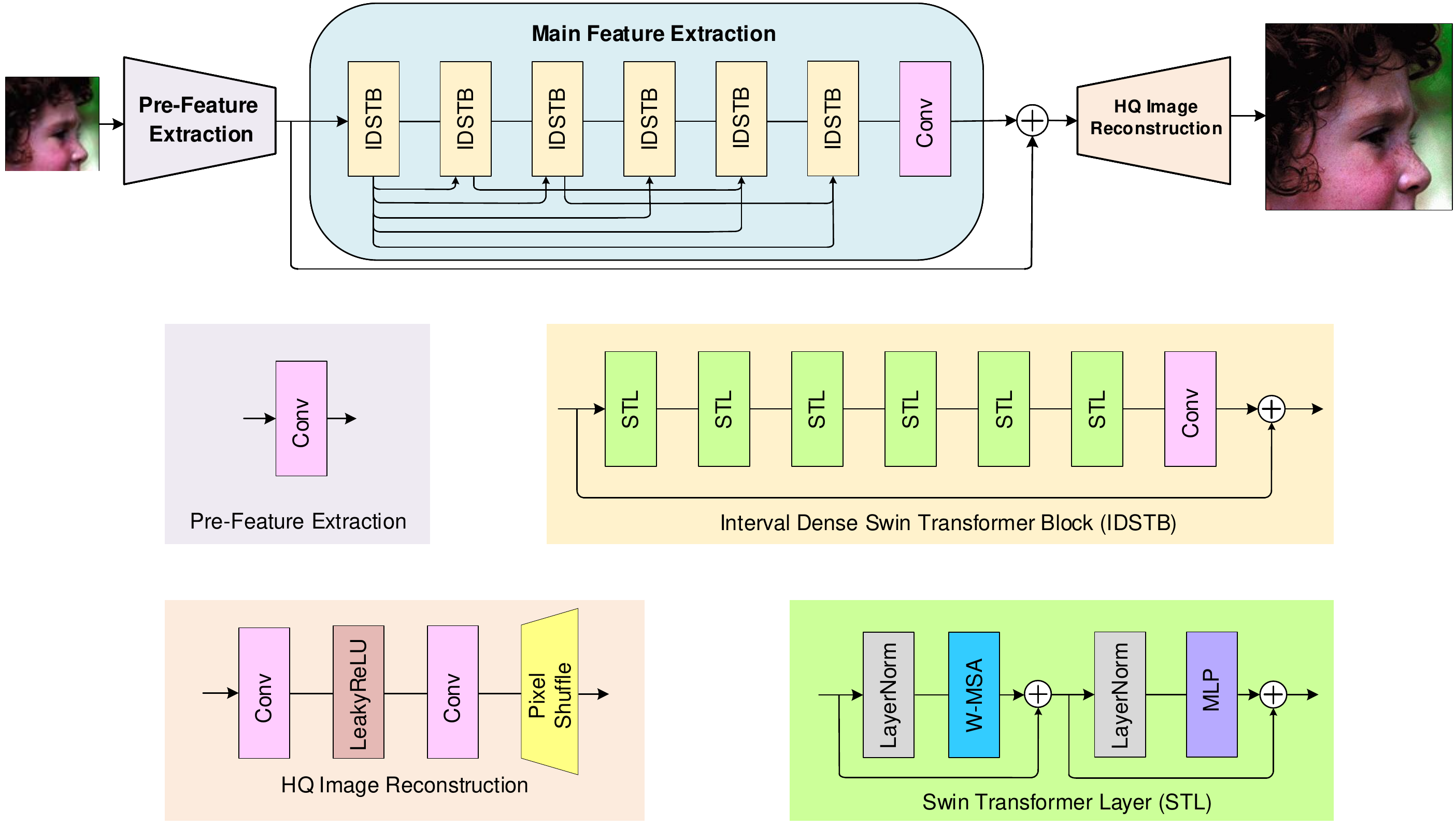}
  \caption{The architecture of our proposed SwinOIR for image super-resolution.}
  \label{fig_swinoir}
\end{figure*}

\section{Related Work}
\label{related}
In recent year, the Transformer model \cite{vaswani2017attention} has demonstrated impressive performance in natural language processing (NLP), and researchers have recently started to apply it to CV tasks. The introduction of ViT (Vision Transformer) \cite{dosovitskiy2020image} has made the self-attention mechanism popular in CV tasks. Unlike traditional CNNs, which learn to identify local patterns and features, Transformers are designed to focus on more important image regions through global interactions by using self-attention mechanisms. This advantage has enabled Transformer models to achieve SOTA performance in classical CV tasks, including image classification \cite{wu2020visual}\cite{vaswani2021scaling}, object detection \cite{carion2020end}\cite{liu2020deep}, and semantic segmentation \cite{cao2021swin}\cite{zheng2021rethinking}. Additionally, researchers have also started to implement Transformers on image restoration tasks \cite{chen2021pre}\cite{wang2022uformer}.

Swin Transformer \cite{liu2021swin} is a general backbone network based on Transformer that can be applied to the classical CV tasks. It divides the feature map into non-overlapping windows of different sizes, and self-attention is computed only within the given window. This model achieves SOTA performance in image classification, object detection, semantic segmentation, and instance segmentation. Therefore, we improve Swin Transformer and apply the newly proposed model to image super-resolution tasks.

SwinIR \cite{liang2021swinir} is an improved network based on Swin Transformer that aims to perform image restoration tasks, such as super-resolution, denoising, and JPEG artifact removal. The framework comprises a set of modules, including the Swin Transformer encoder and decoder, the feature fusion module, and the residual module. The SwinIR encoder and decoder modules are based on the Swin Transformer architecture, which is a hierarchical self-attention network that processes input data in a patch-based method. The feature fusion module is used to merge the features extracted from the encoder and decoder, while the residual module is used to enhance the high-frequency details in the restored image.

Based on SwinIR, Zhang \emph{et al.} \cite{zhang2022swinfir} proposes a Fast Fourier Transform (FFT) convolution layer, which enables the network to process input images in the frequency domain. The improved SwinFIR model allows the network to capture more low-frequency information and better preserve the structural details of the image. For model training, SwinFIR introduces some improvements, such as data augmentation, adversarial training, and progressive learning. For the loss function, the Charbonnier loss \cite{lai2018fast} is used for this model, which is defined as:

\begin{equation}
L(\theta)=\frac{1}{N} \sum_{i=1}^{N} \sqrt{\left(Model\left(I_{L}^{i}, \theta\right)-I_{H}^{i}\right)^{2}+\varepsilon}
\end{equation} 
where $N$ denotes the numbers of training images, and $\theta$ presents the parameters of the network model. These modifications help to reduce overfitting and improve the generalization ability of the network.

On the basis of these models, we apply the Interval Dense Connection Strategy to SwinIR, which enables the newly proposed model to achieve SOTA performance. In Section \ref{ablation}, we conduct an ablation study to demonstrate the positive effect of this strategy on our proposed model.

\section{Proposed Method}
\label{method}
\subsection{SwinOIR Architecture}
For image super-resolution, we modify SwinIR \cite{liang2021swinir} and design the Interval Dense Connection Strategy to connect different blocks. The newly proposed network architecture (SwinOIR) consists of three modules: Pre-Feature Extraction, Main Feature Extraction, and High Quality Image Reconstruction, as illustrated in Fig. \ref{fig_swinoir}.

\subsubsection{Feature Extraction}
In this work, we divide the feature extraction process into two stages: Pre-Feature Extraction and Main Feature Extraction. The previous work by Tete \emph{et al.} \cite{xiao2021early} demonstrated that convolution is suitable for early visual processing. Therefore, we use a $3\times 3$ convolution layer, denoted by $H_{\textit{pre}}(\cdot)$, for the pre-feature extraction stage, resulting in a feature map $F_{pre}\in\mathbb{R}^{H\times W\times C}$, where $H$, $W$, and $C$ represent the height, width, and channel number of the input object image $I_{\textit{O}}\in\mathbb{R}^{H\times W\times C}$, respectively. The equation is given by:

\begin{equation}
F_{pre}=H_{\textit{pre}}(I_{\textit{O}}).
\end{equation} 

After completing pre-feature extraction, this work extracts main feature $F_{\textit{main}}$ extraction from $F_{\textit{pre}}$:

\begin{equation}
F_{\textit{main}}=H_{\textit{main}}(F_{pre}),
\end{equation} 
where $H_{\textit{main}}(\cdot)$ contains $m$ Interval Dense Swin Transformer Blocks (IDSTB) and a $3 \times 3$ convolution layer. Specifically, $m$ is the number of blocks in Main Feature Extraction. It is worth noting that SwinIR \cite{liang2021swinir} uses 6 blocks to form the network architecture, but the goal of this paper is to optimize the model for mobile devices. Therefore we set $m=4$ to reduce the overall number of parameters for deploying the network model on the low computing power platform. For more details about the design of IDSTB, please refer to Section \ref{MFE}.

\subsubsection{Image Reconstruction}
After completing the feature extraction process, the high quality object image is reconstructed using the image reconstruction module $H_{\textit{IR}}(\cdot)$, resulting in an image $I_{\textit{HQO}}\in\mathbb{R}^{H\times W\times C}$. In order to ensure that low-frequency and high-frequency information is properly incorporated, a skip connection is employed to add the information from pre-feature extraction and main feature extraction. The resulting equation is as follows:

\begin{equation}
I_{\textit{HQO}}=H_{\textit{IR}}(F_{\textit{pre}}+F_{\textit{main}}).
\end{equation}

\subsubsection{Loss Function}
In this work, the parameters of the network architecture are optimized using the minimization of $L1$ pixel loss:

\begin{equation}
\mathcal{L}=\| I_{\textit{HQO}}-I_{\textit{HQ}}\|_1,
\end{equation}
where $I_{\textit{HQO}}$ is the image obtained from the input object image by SwinOIR, and $I_{\textit{HQ}}$ is the corresponding high quality ground truth image.

For the classical image super-resolution task, we follow the previous work to evaluate the performance of our network model by using only the $L1$ pixel loss. For real-life super-resolution tasks, this work combines perceptual loss, GAN loss, and pixel loss \cite{goodfellow2020generative}\cite{johnson2016perceptual}\cite{wang2021real}\cite{wang2018esrgan}\cite{zhang2021designing} to improve the image quality.

\subsection{Main Feature Extraction}
\label{MFE}
\subsubsection{Swin Transformer Layer}
In SwinOIR, the Swin Transformer Layer (STL) differs from the one employed in the original Swin Transformer \cite{liu2021swin} architecture. Specifically, this work uses the window-based multi-head self-attention (W-MSA) + multilayer perceptron (MLP) approach instead of the shifted window-based multi-head self-attention (SW-MSA) + MLP employed in the original model \cite{liang2021swinir}.

As shown in Fig. \ref{fig_swinoir}, when an input image of size ${H\times W\times C}$ is provided, Swin Transformer divides it into $M\times M$ non-overlapping local windows to reshape the image into $\frac{HW}{M^2}\times M^2\times C$ features, where $\frac{HW}{M^2}$ denotes the total number of windows, and self-attention is computed independently for each window. For feature $x\in\mathbb{R}^{M^2\times C}$ of a local window, the calculations for the query $Q$, key $K$, and value matrices $V$ are given by the following equations:
\begin{equation}
\begin{split}
Q=xP_Q,\\
K=xP_K,\\
V=xP_V.
\end{split}
\end{equation}

For $Q,K,V\in\mathbb{R}^{M^2\times d}$, the equation of the self-attention  mechanism in the local window is:

\begin{equation}
\operatorname{Attention}(Q, K, V)=\operatorname{softmax}(\frac{Q K^{T}}{\sqrt{d_{k}}} + B) V,
\end{equation}
where $B$ is the relative position encoding and it implements the attention function multiple times and concatenates the results as MSA.

\begin{algorithm}
\caption{Interval Dense Connection Strategy}
\label{alg}
\begin{algorithmic}
\Require {Input features $F_1,F_2, \ldots, F_m$, \\
\; \; \; \; \; \; Convolution operation $H_{\textit{conv}_i}$}
\Ensure {Output features $F_{out}=H_{\textit{conv}_i}(F_m)$}
\For{the number of each IDSTB from $1$ to $m$}
\State {$m$-th IDSTB connect to $2^0$-th IDSTB}
\For{$level$ in $2^1$ to $2^5$}
\If{the number of IDSTB is odd}
\If{$n$ is even $\textbf{and}$ $order$ $\leq$ level}
\State {$level$-th connect to $order$-th}
\EndIf
\EndIf
\If{the number of IDSTB is even}
\If{$n$ is odd $\textbf{and}$ $order$ $\leq$ level}
\State {$level$-th connect to $order$-th}
\EndIf
\EndIf
\EndFor
\State{$F_{out}=H_{\textit{conv}}(F_m)$}
\EndFor
\end{algorithmic}
\end{algorithm}

MLP is composed of two fully connected layers with Gaussian Error Linear Units (GELU) \cite{hendrycks2016gaussian} for feature transformation. To improve the model's stability and performance, we apply Layer Normalization (LN) \cite{ba2016layer} before both W-MSA and MLP layers. The overall equation for the Swin Transformer Layer in our model is as follows:
\begin{equation}
\begin{split}
x&=x+\text{W-MSA}(\text{LN}(x)),\\
x&=x+\text{MLP}(\text{LN}(x)).
\end{split}
\end{equation}

\subsubsection{Interval Dense Swin Transformer Block}
As shown in Fig. \ref{fig_swinoir}, each IDSTB consists of multiple STLs and a convolution layer. For the input feature $F_{i,0}$ of the $i$-th IDSTB, the extracted features are $F_{i,1},F_{i,2}, \ldots, F_{i,m}$:

\begin{equation}
F_{i,j}=H_{\textit{STL}_{i,j}}(F_{i,j}-1),\, j=1,2,3,\cdots,L,
\end{equation}
where $H_{\textit{STL}_{i,j}}(\cdot)$ is the $j$-th STL in the $i$-th IDSTB. The output equation for each IDSTB is as follows:

\begin{equation}
F_{i,out}=H_{\textit{conv}_i}(F_{i,m})+F_{i,0},
\end{equation}
where $H_{conv_i}(\cdot)$ is the convolution layer in the $i$-th IDSTB. The advantage of this design is that, while Transformers have the effect of Spatially-Varying Convolution \cite{vaswani2021scaling}\cite{elsayed2020revisiting}, the convolution layer can increase the translational equivalence of SwinOIR.

\subsubsection{Interval Dense Connection Strategy}
The main feature extraction stage of SwinOIR consists of multiple IDSTBs and a convolution layer that extract intermediate features $F_{1},F_2, \ldots, F_n$. To improve feature reuse, based on recent researches \cite{ju2022threshnet}\cite{ju2023efficient}, we propose a new algorithm (Algorithm \ref{alg}) that connects the IDSTBs:

\begin{equation}
F_i=\left\{\begin{array}{c}
H_{\textit{IDSTB}_i}\left(\left[F_1, F_2, F_4, \cdots, F_{n-1}\right]\right),\\
\text { if } n \% 2=1,\\
\\
H_{\textit{ISDTB}_i}\left(\left[F_1, F_3, F_5, \cdots, F_{n-1}\right]\right),\\
\text { if } n \% 2=0,
\end{array}\right.
\end{equation}
where $H_{\textit{IDSTB}_{i}}(\cdot)$ denotes the $i$-th IDSTB. Since convolution has the effect of inductive bias, adding a convolution layer $H_{\textit{conv}}$ at the end of the feature extraction helps to connect pre-feature extraction and main feature extraction:

\begin{equation}
F_{\textit{final}}= H_{\textit{conv}}(F_n).
\end{equation} 

To demonstrate the positive effect of this connection strategy on the model performance, we perform ablation study in Setction \ref{ablation}. 

\section{Experiment}
\label{experiment}
\subsection{Dataset}
\subsubsection{Image Super-Resolution Dataset}
DIV2K (DIVerse 2K resolution high quality images) dataset \cite{Ignatov2018} is a widely-used dataset in the field of image super-resolution. This dataset was originally provided for the 2017 NTIRE competition and consists of 1,000 high-quality images. Among these images, 800 are used for training, 100 for validation, and 100 for testing. The training set includes both high-resolution and corresponding low-resolution images, which is useful for training super-resolution models.

Flickr2K \cite{timofte2017ntire} is a commonly used dataset in image super-resolution. This dataset contains 2,560 diverse images with various objects and detailed patterns. It is suitable for a wide range of experimental scenarios and is often used to demonstrate the ability of models to handle diverse information.

Set5 \cite{bevilacqua2012low} and Set14 \cite{zeyde2012single} are popular choices for testing the performance of image super-resolution models. Set5 includes 5 color images, while Set14 includes 14 color images. These datasets are widely used for testing due to their small size and diverse image content.

BSD100 (Berkeley Segmentation Dataset) \cite{martin2001database} is also commonly used for testing super-resolution models. This dataset includes 100 images of size 320$\times$480 and 480$\times$320, and the test images have various types of spatial distributions of key visual information.

Urban100 dataset \cite{huang2015single} is a collection of 100 images from different urban scenes, which is often used as a test set for evaluating the performance of super-resolution models.

\subsubsection{Object Detection Dataset}
MS COCO (Microsoft Common Objects in Context) dataset \cite{lin2014microsoft} is a benchmark dataset for image recognition provided by the Microsoft team. The COCO 2017 dataset is a large-scale dataset with 80 categories, consisting of 118,287 training images, 5,000 validation images, and 40,670 test images. While only the training and validation images are labeled, the test images lack any label information. Researchers can evaluate their proposed neural networks' performance on this dataset and compare their results to other models.

\subsection{Evaluation Metric}
\subsubsection{Peak Signal-to-Noise Ratio}
Peak Signal-to-Noise Ratio (PSNR) is a widely used metric for evaluating the degree of distortion in an image. The PSNR value measures the similarity between two images and is higher when the quality of the distorted image is closer to the original image. It is defined as:

\begin{equation}
PSNR=10 \log _{10}\left(\frac{MAX_{I}^{2}}{MSE}\right),
\end{equation}
where $MSE$ is the mean square error of the two images, and $MAX_I$ is the maximum value of the image pixels. Generally, when the PSNR value exceeds 28 dB, the human eye cannot distinguish the difference between the original and the distorted images. When the PSNR value exceeds 30 dB, the quality of the image is considered good.

\subsubsection{Structural Similarity}
Structural Similarity (SSIM) Index is a widely used metric for measuring the similarity between two images in terms of their luminance, contrast, and structure, and is calculated using the following equation:

\begin{equation}
SSIM(x,y) = \frac{(2\mu_x\mu_y+c_1)(2\sigma_{xy}+c_2)}{(\mu_x^2+\mu_y^2+c_1)(\sigma_x^2+\sigma_y^2+c_2)},
\end{equation}
where $\mu_x$ and $\mu_y$ are the average grayscale values of the images $x$ and $y$, respectively; $\sigma_x$ and $\sigma_y$ are the standard deviations of $x$ and $y$, and $\sigma_{xy}$ is their covariance. The constants $c_1$ and $c_2$ are used to avoid instability when the denominator approaches zero.

The SSIM index ranges between -1 and 1, with 1 indicating perfect similarity between the two images. Typically, an SSIM value greater than 0.9 is considered to indicate very high similarity, while a value less than 0.7 indicates significant differences between the two images.

\begin{table*}
\begin{center}
\caption{Ablation study of the connection method between different blocks in Swin Transformer for lightweight image super-resolution.}
\label{tab_ablation}
\setlength{\tabcolsep}{0.6mm}{
\begin{tabular}{cccccccccccc}
\hline
\multirow{2}{*}{\textbf{Method}} & \multirow{2}{*}{\textbf{Scale}} & \multirow{2}{*}{\textbf{\begin{tabular}[c]{@{}c@{}}Number\\ Block\end{tabular}}} & \multirow{2}{*}{\textbf{\begin{tabular}[c]{@{}c@{}}Connecion\\ Method\end{tabular}}} & \multicolumn{2}{c}{\textbf{Set5}\cite{bevilacqua2012low}} & \multicolumn{2}{c}{\textbf{Set14}\cite{zeyde2012single}} & \multicolumn{2}{c}{\textbf{BSD100}\cite{martin2001database}} & \multicolumn{2}{c}{\textbf{Urban100}\cite{huang2015single}} \\
 &  &  &  & \textbf{PSNR} & \textbf{SSIM} & \textbf{PSNR} & \textbf{SSIM} & \textbf{PSNR} & \textbf{SSIM} & \textbf{PSNR} & \textbf{SSIM} \\ \hline
Baseline & $\times$2 & \begin{tabular}[c]{@{}c@{}}4\\ RSTB\end{tabular} & \begin{tabular}[c]{@{}c@{}}Skip\\ Connection\end{tabular} & 38.12 & 0.960 & 33.83 & 0.920 & 32.26 & 0.900 & 32.58 & 0.932 \\ \hdashline
Ours & $\times$2 & \begin{tabular}[c]{@{}c@{}}4\\ IDSTB\end{tabular} & \begin{tabular}[c]{@{}c@{}}Interval Dense\\ Connection\end{tabular} & 38.14 & 0.961 & 33.86 & 0.921 & 32.27 & 0.901 & 32.71 & 0.932 \\ \hline
Baseline & $\times$3 & \begin{tabular}[c]{@{}c@{}}4\\ RSTB\end{tabular} & \begin{tabular}[c]{@{}c@{}}Skip\\ Connection\end{tabular} & 34.56 & 0.928 & 30.53 & 0.846 & 29.19 & 0.808 & 28.57 & 0.862 \\ \hdashline
Ours & $\times$3 & \begin{tabular}[c]{@{}c@{}}4\\ IDSTB\end{tabular} & \begin{tabular}[c]{@{}c@{}}Interval Dense\\ Connection\end{tabular} & 34.56 & 0.929 & 30.56 & 0.847 & 29.21 & 0.809 & 28.65 & 0.863  \\ \hline
Baseline & $\times$4 & \begin{tabular}[c]{@{}c@{}}4\\ RSTB\end{tabular} & \begin{tabular}[c]{@{}c@{}}Skip\\ Connection\end{tabular} & 32.40 & 0.896 & 28.79 & 0.786 & 27.68 & 0.741 & 26.43 & 0.797 \\ \hdashline
Ours & $\times$4 & \begin{tabular}[c]{@{}c@{}}4\\ IDSTB\end{tabular} & \begin{tabular}[c]{@{}c@{}}Interval Dense\\ Connection\end{tabular} & 32.49 & 0.898 & 28.89 & 0.788 & 27.74 & 0.743 & 26.62 & 0.803 \\ \hline
\end{tabular}}
\end{center}
\end{table*}

\subsubsection{Mean Average Precision}
Mean Average Precision (mAP) is a metric for evaluating models in object detection tasks, and the equation is shown below:

\begin{equation}
mAP = \frac{1}{n_c} \sum_{i=1}^{n_c} AP_i,
\end{equation}
where $n_c$ is the number of object categories, and $AP_i$ is the average precision (AP) for category $i$. AP is calculated by first computing precision and recall values at different confidence thresholds, and then integrating the precision-recall curve to obtain the AP value. Precision and recall are important measures for evaluating the performance of object detection models. Precision is defined as the ratio of true positives (TP) to the total number of predicted positives (TP+FP):

\begin{equation}
Precision = \frac{TP}{TP + FP}.
\end{equation}
Recall is defined as the ratio of true positives (TP) to the total number of actual positives (TP+FN):

\begin{equation}
Recall = \frac{TP}{TP + FN}.
\end{equation}

Intersection over Union (IoU) is used to measure the overlap between the predicted bounding box and the ground truth bounding box. In the MS COCO dataset, mAP 0.5:0.95 means the average precision across all object categories and IoU thresholds ranging from 0.5 to 0.95.

\subsection{Experiment Setup}
\label{setup}
We train the SwinOIR model on DIV2K \cite{Ignatov2018} and DF2K (DIV2K + Flickr2K \cite{timofte2017ntire}) datasets and evaluate on Set5, Set14, BSD100, and Urban100 benchmarks. The model is configured with an IDSTB number of 4, a STL number of 4, a window size of 8, a channel number of 60, and an attention head number of 6.

This work employes the AdamW optimizer \cite{loshchilov2018decoupled} with $\beta_1 = 0.9$ and $\beta_2 = 0.9$, and the learning rate is initialized to $5 \times 10^{-4}$. The learning rate is reduced by half at 300, 600, and 900 epochs. The model is trained for 1,000 epochs using Pytorch 1.9.0 on a single GPU NVIDIA GeForce RTX 3090. For $\times$2, $\times$3, and $\times$4 upscaling image super-resolution tasks, we use batch sizes of 64, 48, and 24, respectively.

\subsection{Ablation Study}
\label{ablation}
In this section, this work conducts an ablation study to demonstrate the contribution of the Interval Dense Connection Strategy to the model. It uses SwinIR \cite{liang2021swinir} as the baseline model, which uses skip connection to connect between different blocks, and our model named ``ours'', which uses the interval dense connection to connect between different blocks. To ensure that the model performance is not affected by other model parameters, we set the number of blocks for the baseline model and our model to 4, and the number of STLs of each block to 4. This work trains all models on the DIV2K dataset \cite{Ignatov2018}, and evaluates the model performances for $\times$2, $\times$3, and $\times$4 upscaling image super-resolution on Set5 \cite{bevilacqua2012low}, Set14 \cite{zeyde2012single}, BSD100 \cite{martin2001database}, and Urban100 \cite{huang2015single} datasets.

As can be seen from Table \ref{tab_ablation}, the PSNR and SSIM of our model are mostly higher than those of the baseline model on all benchmark datasets, except for the PSNR of $\times$3 upscaling image super-resolution on Set5 dataset, and the SSIM of $\times$2 upscaling image super-resolution on urban100 dataset,  which are equal. The PSNRs of our model on Urban100 dataset at different scales have shown significant improvements compared to the baseline model. Specifically, our model's PSNRs have improved by 0.13 dB, 0.08 dB, and 0.19 dB at the scales of $\times$2, $\times$3, and $\times$4, from 32.58 dB, 28.57 dB, and 26.43 dB to 32.71 dB, 28.65 dB, and 26.62 dB, respectively. The results of ablation study demonstrate that the Interval Dense Connection Strategy has a more positive effect on the model than the skip connection. The improvements obtained by our model indicate the effectiveness of our proposed strategy in enhancing the quality of the low-quality images.

\begin{figure*}[h]
  \centering
  \includegraphics[width=\linewidth]{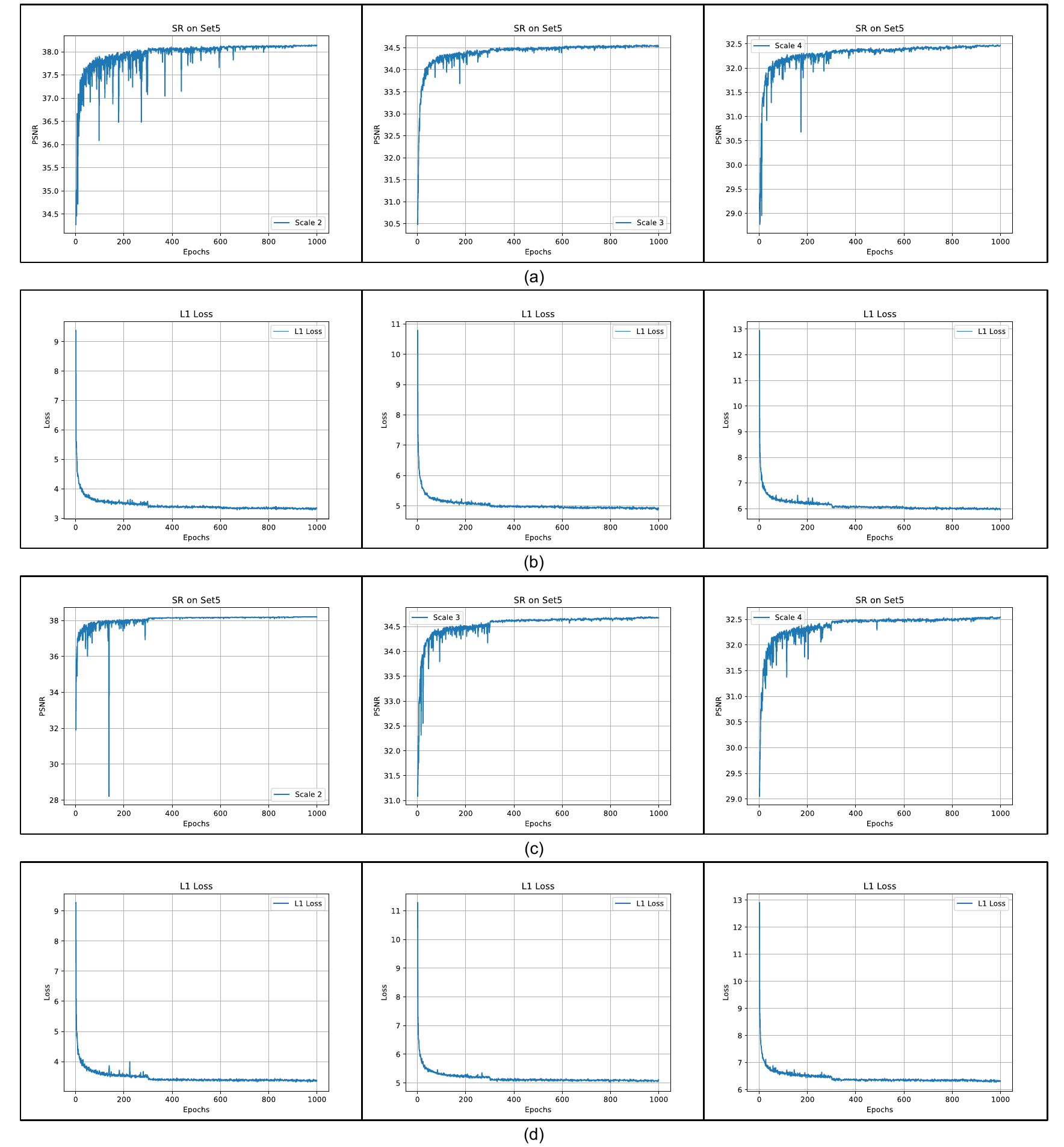}
  \caption{Visualization information of SwinOIR model training for $\times$2, $\times$3 and $\times$4 upscaling image super-resolution. (a) The curve chart of epochs and PSNR trained on DIV2K dataset, (b) the curve chart of epochs and loss trained on DIV2K dataset, (c) the curve chart of epochs and PSNR trained on DF2K dataset, (d) the curve chart of epochs and loss trained on DF2K dataset.}
  \label{fig_loss}
\end{figure*}

\begin{table*}
\centering
\caption{Quantitative comparison (average PSNR/SSIM) with other state-of-the-art methods/models for lightweight image super-resolution on benchmark datasets. Best and 2nd best performance are in {\color{red}red} and {\color{blue}blue} colors, respectively.}
\label{tab_comparison}
\setlength{\tabcolsep}{1.2mm}{
\begin{tabular}{ccccccccccc}
\hline
\multirow{2}{*}{\textbf{Method}} & \multirow{2}{*}{\textbf{Scale}} & \multirow{2}{*}{\textbf{\begin{tabular}[c]{@{}c@{}}\#Params\\ (K)\end{tabular}}} & \multicolumn{2}{c}{\textbf{Set5}\cite{bevilacqua2012low}} & \multicolumn{2}{c}{\textbf{Set14}\cite{zeyde2012single}} & \multicolumn{2}{c}{\textbf{BSD100}\cite{martin2001database}} & \multicolumn{2}{c}{\textbf{Urban100}\cite{huang2015single}} \\
 &  &  & \textbf{PSNR} & \textbf{SSIM} & \textbf{PSNR} & \textbf{SSIM} & \textbf{PSNR} & \textbf{SSIM} & \textbf{PSNR} & \textbf{SSIM} \\
\hline
IMDN\cite{hui2019lightweight} & $\times$2 & 694 & 38.00 & 0.9605 & 33.63 & 0.9177 & 32.19 & 0.8996 & 32.17 & 0.9283 \\
LAPAR-A\cite{li2020lapar} & $\times$2 & 548 & 38.01 & 0.9605 & 33.62 & 0.9183 & 32.19 & 0.8999 & 32.10 & 0.9283 \\
LatticeNet\cite{luo2020latticenet} & $\times$2 & 756 & {\color{blue}38.15} & 0.9610 & 33.78 & 0.9193 & 32.25 & 0.9005 & 32.43 & 0.9302 \\
SwinIR\cite{liang2021swinir} & $\times$2 & 878 & 38.14 & 0.9611 & {\color{blue}33.86} & 0.9206 & {\color{blue}32.31} & 0.9012 & {\color{blue}32.76} & {\color{blue}0.9340} \\
CARN\cite{ahn2018fast} & $\times$2 & 1,592 & 37.76 & 0.9590 & 33.52 & 0.9166 & 32.09 & 0.8978 & 31.92 & 0.9256 \\
FALSR-A\cite{chu2021fast} & $\times$2 & 1,021 & 37.82 & 0.9590 & 33.55 & 0.9168 & 32.10 & 0.8987 & 31.93 & 0.9256 \\
\textbf{Ours} & $\times$2 & 841 & 38.14 & {\color{blue}0.9612} & {\color{blue}33.86} & {\color{blue}0.9208} & 32.27 & {\color{blue}0.9014} & 32.71 & 0.9328 \\
\textbf{Ours+} & $\times$2 & 841 & {\color{red}38.21} & {\color{red}0.9614} & {\color{red}33.97} & {\color{red}0.9220} & {\color{red}32.34} & {\color{red}0.9022} & {\color{red}32.83} & {\color{red}0.9353} \\ \hline
IMDN\cite{hui2019lightweight} & $\times$3 & 703 & 34.36 & 0.9270 & 30.32 & 0.8417 & 29.09 & 0.8046 & 28.17 & 0.8519 \\
LAPAR-A\cite{li2020lapar} & $\times$3 & 544 & 34.36 & 0.9267 & 30.34 & 0.8421 & 29.11 & 0.8054 & 28.15 & 0.8523 \\
LatticeNet\cite{luo2020latticenet} & $\times$3 & 765 & 34.53 & 0.9281 & 30.39 & 0.8424 & 29.15 & 0.8059 & 28.33 & 0.8538 \\
SwinIR\cite{liang2021swinir} & $\times$3 & 886 & {\color{blue}34.62} & 0.9289 & 30.54 & 0.8463 & 29.20 & 0.8082 & {\color{blue}28.66} & 0.8624 \\
CARN\cite{ahn2018fast} & $\times$3 & 1,592 & 34.29 & 0.9255 & 30.29 & 0.8407 & 29.06 & 0.8034 & 28.06 & 0.8493 \\
\textbf{Ours} & $\times$3 & 1,025 & 34.56 & {\color{blue}0.9290} & {\color{blue}30.56} & {\color{blue}0.8470} & {\color{blue}29.21} & {\color{blue}0.8090} & 28.65 & {\color{blue}0.8626} \\
\textbf{Ours+} & $\times$3 & 1,025 & {\color{red}34.69} & {\color{red}0.9296} & {\color{red}30.65} & {\color{red}0.8493} & {\color{red}29.27} & {\color{red}0.8111} & {\color{red}28.87} & {\color{red}0.8674} \\ \hline
IMDN\cite{hui2019lightweight} & $\times$4 & 715 & 32.21 & 0.8948 & 28.58 & 0.7811 & 27.56 & 0.7353 & 26.04 & 0.7838 \\
LAPAR-A\cite{li2020lapar} & $\times$4 & 659 & 32.15 & 0.8944 & 28.61 & 0.7818 & 27.61 & 0.7366 & 26.14 & 0.7871 \\
LatticeNet\cite{luo2020latticenet} & $\times$4 & 777 & 32.30 & 0.8962 & 28.68 & 0.7830 & 27.62 & 0.7367 & 26.25 & 0.7873 \\
SwinIR\cite{liang2021swinir} & $\times$4 & 897 & 32.44 & 0.8976 & 28.77 & 0.7858 & 27.69 & 0.7406 & 26.47 & 0.7980 \\
CARN\cite{ahn2018fast} & $\times$4 & 1,592 & 32.13 & 0.8937 & 28.60 & 0.7806 & 27.58 & 0.7349 & 26.07 & 0.7837 \\
\textbf{Ours} & $\times$4 & 988 & {\color{blue}32.49} & {\color{blue}0.8978} & {\color{blue}28.89} & {\color{blue}0.7880} & {\color{blue}27.74} & {\color{blue}0.7430} & {\color{blue}26.62} & {\color{blue}0.8031} \\
\textbf{Ours+} & $\times$4 & 988 & {\color{red}32.55} & {\color{red}0.8980} & {\color{red}28.92} & {\color{red}0.7892} & {\color{red}27.76} & {\color{red}0.7441} & {\color{red}26.74} & {\color{red}0.8060} \\ \hline
\end{tabular}}
\end{table*}

\begin{table*}[]
\centering
\caption{Model performance comparison of YOLO algorithms on the edge computing platform and CPU, and the model performance improvement and performance degradation are in {\color{red}red} and {\color{blue}blue}.}
\label{detecion_model}
\setlength{\tabcolsep}{2.2mm}{
\begin{tabular}{ccccccc}
\hline
\textbf{Model} & \textbf{\begin{tabular}[c]{@{}c@{}}$\rm \bf mAP^{val}$\\ 50\end{tabular}} & \textbf{\begin{tabular}[c]{@{}c@{}}$\rm \bf mAP^{val}$\\ 50-95\end{tabular}} & \textbf{\begin{tabular}[c]{@{}c@{}}FPS\tnote{1}\\ Jetson AGX Orin\end{tabular}} & \textbf{\begin{tabular}[c]{@{}c@{}}Speed\\ CPU ONNX\tnote{2}\\ (ms)\end{tabular}} & \textbf{\begin{tabular}[c]{@{}c@{}}\#Params\\ (M)\end{tabular}} & \textbf{\begin{tabular}[c]{@{}c@{}}FLOPs\\ (B)\end{tabular}} \\ \hline
YOLOv5n & 28.0 & 45.7 & 370 & 45 & 1.9 & 4.5 \\
YOLOv8n & 37.3 & 52.5 & 383 & 80 & 3.2 & 8.7 \\
Difference & {\color{red}33.2\%} & {\color{red}14.9\%} & {\color{blue}3.5\%} & {\color{blue}77.8\%} & {\color{blue}68.4\%} & {\color{blue}93.3\%} \\ \hline
YOLOv5s & 37.4 & 56.8 & 277 & 98 & 7.2 & 16.5 \\
YOLOv8s & 44.9 & 61.8 & 260 & 128 & 11.2 & 28.6 \\
Difference & {\color{red}20.1\%} & {\color{red}8.8\%} & {\color{red}6.1\%} & {\color{blue}30.6\%} & {\color{blue}55.6\%} & {\color{blue}73.3\%} \\ \hline
YOLOv5m & 45.4 & 64.1 & 160 & 224 & 21.2 & 49.0 \\
YOLOv8m & 50.2 & 67.2 & 137 & 235 & 25.9 & 78.9 \\
Difference & {\color{red}10.6\%} & {\color{red}4.8\%} & {\color{red}14.4\%} & {\color{blue}4.9\%} & {\color{blue}22.1\%} & {\color{blue}61.0\%} \\ \hline
YOLOv5l & 49.0 & 67.3 & 116 & 430 & 46.5 & 109.1 \\
YOLOv8l & 52.9 & 69.8 & 95 & 375 & 43.7 & 165.2 \\
Difference & {\color{red}8.0\%} & {\color{red}3.7\%} & {\color{red}18.1\%} & {\color{red}12.8\%} & {\color{red}6.0\%} & {\color{blue}51.4\%} \\ \hline
YOLOv5x & 50.7 & 68.9 & 67 & 766 & 86.7 & 205.7 \\
YOLOv8x & 53.9 & 71.0 & 64 & 479 & 68.2 & 257.8 \\
Difference & {\color{red}6.3\%} & {\color{red}3.0\%} & {\color{red}4.5\%} & {\color{red}37.5\%} & {\color{red}21.3\%} & {\color{blue}25.3\%} \\ \hline
\end{tabular}}
\begin{tablenotes}
\footnotesize
\item[$^1$]
FPS Jeston AGX Orin is the test result of frame per second (FPS) for object detection tasks performed on the models using NVIDIA JetPack 5.0 development kit on NVIDIA Jetson AGX Orin.
\item[$^2$]
Speed CPU ONNX is the total inference time per image for object detection tasks performed on the models in onnx format on CPU Intel Core.
\end{tablenotes}
\end{table*}

\begin{figure*}[h]
  \centering
  \includegraphics[width=\linewidth]{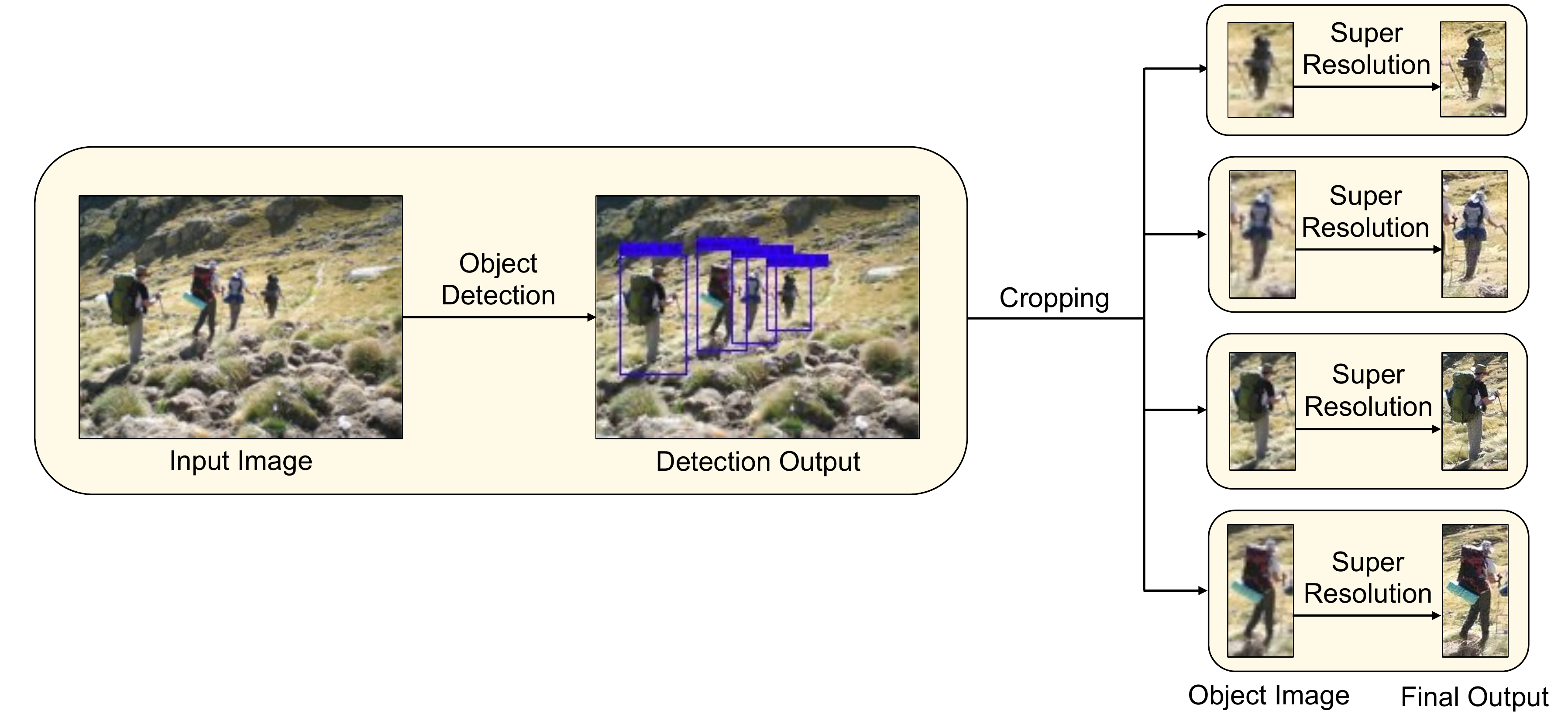}
  \caption{Overall flowchart of the two-stage framework for the real-life application.}
  \label{fig_stage}
\end{figure*}

\begin{figure*}[h]
  \centering
  \includegraphics[width=\linewidth]{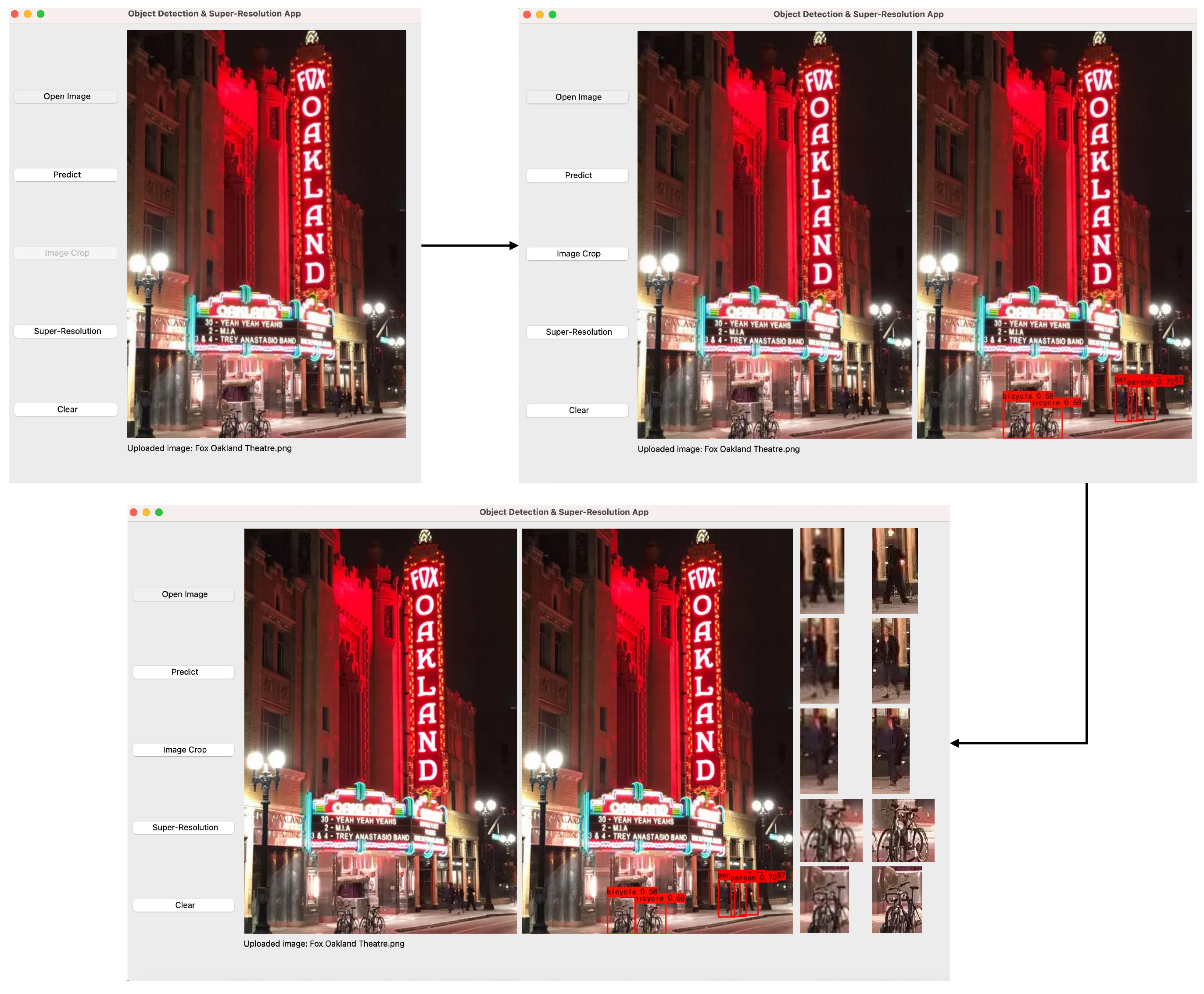}
  \caption{Example of the using flow of our designed application on macOS.}
  \label{fig_app}
\end{figure*}

\subsection{Experimental Results}
To depict the training process of SwinOIR model in greater detail, we generate curve charts of PSNR, Loss, and Epochs. Each row from left to right of Fig. \ref{fig_loss} displays the visualization information of SwinOIR model training for $\times$2, $\times$3, and $\times$4 upscaling image super-resolution, respectively. This work illustrates the training process of our models on DIV2K \cite{Ignatov2018} and DF2K \cite{timofte2017ntire} datasets, corresponding to "Ours" and "Ours+" models in Table \ref{tab_comparison}, respectively. The curves in Fig. \ref{fig_loss} demonstrate a substantial improvement in training up to 300 and 600 epochs, as the learning rate decays by half at this point. According to the report \cite{liang2021swinir}, a larger initial learning rate can accelerate training, while decaying the learning rate during training can enable the model to be more stable. Therefore, we utilize this warm-up strategy when training SwinOIR, and the graphs demonstrate that this training method is effective in improving training results and performance.

To evaluate the performance of our model, we compare the proposed SwinOIR model with other SOTA lightweight image super-resolution models, such as IMDN \cite{hui2019lightweight}, LAPAR \cite{li2020lapar}, LatticeNet \cite{luo2020latticenet}, CARN \cite{ahn2018fast}, FALSR \cite{chu2021fast}, and SwinIR \cite{liang2021swinir}. Since the focus of this work is on applying the image super-resolution models to low computing power platforms, it evaluates the model performance in terms of PSNR, SSIM, and the number of model parameters, to evaluate the size of different models. As shown in Table \ref{tab_comparison}, we add data to optimize the performance of our model, named ``Ours+''. Specifically, this work expands DIV2K dataset \cite{Ignatov2018} to DF2K dataset \cite{timofte2017ntire}, increasing the number of images in the training set.

Table \ref{tab_comparison} presents the quantitative results of image super-resolution for $\times$2, $\times$3, and $\times$4 upscaling on multiple benchmark datasets. Compared with other SOTA lightweight image super-resolution models, this work achieves the best super-resolution performance at the $\times$2, $\times$3, and $\times$4 scales using the optimized model with increasing data. SwinOIR achieves the top two best model performance on Set5 \cite{bevilacqua2012low}, Set14 \cite{zeyde2012single}, BSD100 \cite{martin2001database}, and Urban100 \cite{huang2015single} datasets without using the increasing data optimization. In particular, SwinOIR model outperforms SwinIR model in terms of PSNR on the Set14 and Urban100 datasets at the $\times$4 scale, from 28.77 dB and 26.47 dB to 28.89 dB and 26.62 dB, with improvements of 0.12 dB and 0.15dB, respectively. This improvement demonstrates the effectiveness of our interval dense connection strategy, which connects different blocks of SwinOIR.

\section{Application}
\label{application}
In this paper, we propose a framework that combines SwinOIR model with You Only Look Once (YOLO) algorithm for real-life scenarios applications. As depicted in Fig. \ref{fig_stage}, our framework comprises two stages, where the first stage uses the YOLO algorithm model to perform object detection on input images, and the second stage performs resolution enhancement on the cropped images. Generally, the first stage produces multiple object images, which are parts of the original input image. Our work applies super-resolution processing to each object image separately, allowing multiple object images easier to be identified, and enhancing object details recognition.

The Ultralytics HUB is a powerful machine learning and deployment platform, and the Ultralytics team has proposed the YOLOv5 \cite{glenn2022} and YOLOv8 \cite{glenn2023} algorithms in 2021 and 2023, respectively. Both of these algorithms aim to perform pattern recognition on mobile devices. In our work, we evaluated the performance of the two algorithms on Jetson AGX Orin and CPU, and the experimental results are shown in Table \ref{detecion_model}. Compared with YOLOv5, YOLOv8 has a higher accuracy rate, up to 33.2\%. However, due to the larger parameter size of YOLOv8 algorithm models, the inference time on CPU is longer. Therefore, it is more reasonable to combine the YOLOv8 algorithm model with the strong computing power platforms.

After completing our SwinOIR model training and YOLOv8 model comparison, we export these two models to the ONNX format and use PySide6 to create a Graphical User Interface (GUI) application, named ``Object Detection \& Super-Resolution App'', as shown in Fig. \ref{fig_app}. In this application, users can open an image, perform object detection, crop the object image, and finally perform super-resolution processing on the object image to identify tiny details in the low-quality image.

\section{Conclusions and Future Work}
\label{conclusion}
In this paper, we present SwinOIR, an image super-restoration model that addresses the limitations of SwinIR feature weight iterative update capability by implementing the Interval Dense Connection Strategy. This method improves the model performance by enabling interval dense connection between different blocks, leading to the SOTA performance on the multiple popular image super-resolution benchmark datasets.

This work has demonstrated the SOTA performance of SwinOIR for image super-resolution tasks. In future research, we will apply SwinOIR to other image restoration tasks, such as image denoising, image deraining, defocus deblurring, image motion deblurring, and JPEG compression artifact reduction.

This paper also demonstrates the practical utility of SwinOIR model by integrating it with the YOLOv8 object detection algorithm to create an application for real-life scenarios. This application is currently available on the macOS and can be run on both CPU and GPU. In the future, we plan to extend the availability of this application to the iOS and Android, enabling users to run it on their mobile devices.

\section{Declarations}
\subsection{Funding}
This research work was supported in part by the National Science and Technology Council, Taiwan, under grant number: NSTC 111-2221-E032-021-.

\subsection{Competing interests}
The authors have no financial or proprietary interests in any material discussed in this article.

\subsection{Ethics approval}
This research does not involve human participants and/or animals.

\bibliographystyle{spmpsci}
\bibliography{reference}

\end{document}